\documentclass[pmlr]{jmlr}

\makeatletter
  \let\Ginclude@graphics\@org@Ginclude@graphics
\makeatother

\usepackage{longtable}

\usepackage{multibib}
\newcites{appendix}{References}

\usepackage{silence}
\usepackage{float}

\usepackage{booktabs}

\usepackage{comment}   
\usepackage{chngcntr}  
\usepackage{booktabs}  
\usepackage{mathtools} 
\usepackage{multirow}  
\usepackage{paralist}  
\usepackage{url}       

\usepackage[load-configurations=version-1]{siunitx}

\DeclareMathOperator*{\argmin}{arg\,min}
\DeclareMathOperator*{\E}{\mathbb{E}}

\usepackage{bm}      
\usepackage{amssymb} 

\usepackage{tikz}
\usetikzlibrary{arrows, decorations.markings, chains}
\usepackage{adjustbox}
\usepackage{pgfplots} 
\usepgfplotslibrary{fillbetween}
\usepgfplotslibrary{groupplots}
\pgfplotsset{compat=1.15}

\tikzstyle{vecArrow} = [thick, decoration={markings,mark=at position
	1 with {\arrow[semithick]{open triangle 60}}},
double distance=1.4pt, shorten >= 5.5pt,
preaction = {decorate},
postaction = {draw,line width=1.4pt, white,shorten >= 4.5pt}]
\tikzstyle{innerWhite} = [semithick, white,line width=1.4pt, shorten >= 4.5pt]
\tikzstyle{line} = [draw, thick, black!50!green, -latex']

\newcommand{\FTB}[1]{}

\jmlrvolume{106}
\jmlryear{2019}
\jmlrworkshop{Machine Learning for Healthcare}

\title[Early Recognition of Sepsis with MGP-TCNs]{Early Recognition of Sepsis with Gaussian Process Temporal Convolutional Networks and Dynamic Time Warping}

\author{%
    \Name{Michael Moor}\footnotemark[1]\textsuperscript{,}\footnotemark[2]         \Email{michael.moor@bsse.ethz.ch}\\
    \Name{Max Horn}\footnotemark[1]\textsuperscript{,}\footnotemark[2]         \Email{max.horn@bsse.ethz.ch}\\ 
    \Name{Bastian Rieck}\footnotemark[1]\textsuperscript{,}\footnotemark[2]     \Email{bastian.rieck@bsse.ethz.ch}\\
    \Name{Damian Roqueiro}\footnotemark[1]\textsuperscript{,}\footnotemark[2]   \Email{damian.roqueiro@bsse.ethz.ch}\\
    \Name{Karsten Borgwardt}\footnotemark[1]\textsuperscript{,}\footnotemark[2] \Email{karsten.borgwardt@bsse.ethz.ch}%
    \AND
    \addr \footnotemark[1]Department of Biosystems Science and Engineering, ETH Zurich, Switzerland\\
    \addr \footnotemark[2]SIB Swiss Institute of Bioinformatics, Switzerland
}

\begin{document}

\maketitle

\begin{abstract}
Sepsis is a life-threatening host response to infection that is associated with high mortality, morbidity, and health costs. Its management is highly time-sensitive because each hour of
delayed treatment increases mortality due to irreversible organ damage.
Meanwhile, despite decades of clinical research, robust biomarkers for
sepsis are missing. Therefore, detecting sepsis early by utilizing the
affluence of high-resolution intensive care records has become
a challenging machine learning problem. Recent advances in deep learning
and data mining promise to deliver a powerful set of tools to efficiently address this task.
This empirical study proposes two novel approaches for the early
detection of sepsis: a deep learning model and a lazy learner that is based on
time series distances.  Our deep learning model employs a temporal
convolutional network that is embedded in a multi-task Gaussian Process
adapter framework, making it directly applicable to irregularly-spaced
time series data.
In contrast, our lazy learner is an ensemble approach that employs
dynamic time warping.
We frame the timely detection of sepsis as a supervised time series
classification task. Consequently, we derive the most recent sepsis
definition in an hourly resolution to provide the first fully
accessible early sepsis detection environment. Seven hours before
sepsis onset, our methods improve area under the precision--recall
curve from 0.25 to 0.35 and 0.40, respectively, over the state of the art. This
demonstrates that they are well-suited for detecting sepsis in the
crucial earlier stages when management is most effective.
\end{abstract}

\section{Introduction}
\label{intro}

Sepsis is defined as a life-threatening organ
dysfunction that is caused by a dysregulated host response to infection \citep{singer2016third}. Despite decades
of clinical research, sepsis remains a major public health issue that is associated with
high mortality, morbidity, and related health
costs~\citep{dellinger2013, kaukonen2014, hotchkiss2016}.

Currently, when sepsis is detected and the underlying pathogen is
identified, organ damage has already progressed to a potentially
irreversible stage. Effective management, especially in the intensive
care unit~(ICU), is of critical importance. From sepsis onset, each hour of delayed effective
antibiotic treatment increases mortality~\citep{ferrer2014empiric}.
Therefore, early detection of sepsis has gained considerable attention in the machine learning
community \citep{kam2017learning,
futoma2017learning}.
The task of detecting sepsis early has often been modeled as a multi-channel time
series classification task.
Clinical data is commonly sampled irregularly, thus often requiring a set of
hand-crafted preprocessing steps, such as binning, carry-forward
imputation, and rolling means~\citep{calvert2016, desautels2016prediction} prior
to the application of a predictive model.
However, these imputation schemes lead to a loss of data sparsity, which
may carry crucial information in this context. Most existing approaches
are incapable of retaining sampling information, thereby potentially
impeding the training and leading to lower predictive performance.
\citet{futoma2017learning} proposed a sepsis detection method that
accounts for irregular sampling by applying the Gaussian Process adapter
end-to-end learning framework \citep{li2016scalable} and then training it
using a long short-term memory~(LSTM) classifier~\citep{hochreiter1997long}.
Only recently have convolutional networks gained attention in sequence
modeling \citep{gehring2017convolutional, vaswani2017attention}. In
particular, temporal convolutional networks~(TCNs; \citet{lea2017temporal}),
have been shown to outperform conventional recurrent neural network~(RNN) architectures for many sequential learning tasks in terms of evaluation metrics,
memory efficiency, and parallelism~\citep{bai2018empirical}.
In light of these developments, we propose a deep learning model as an
end-to-end trainable framework for early sepsis detection that builds on
both multi-task Gaussian Process~(MGP) adapters~(which are an extension of
Gaussian Process adapters to multi-task learning) and TCNs. We refer to
this model as MGP-TCN because it combines the uncertainty-aware framework of GP
adapters with TCNs.
The contributions of our work are threefold:
\begin{itemize}
\item We present a lazy learner that is based on dynamic time warping and
  $k$-nearest neighbors~(DTW-KNN), which can be seen as a multi-channel
  ensemble extension of a well-established data mining technique for
  time series classification. Moreover, we develop MGP-TCN, which is the first
  model that can leverage temporal convolutions on irregularly-sampled
  multivariate time series.
    \item We provide the first fully-accessible framework for the early
      detection of sepsis on a benchmark dataset featuring a publicly
      available temporally resolved Sepsis-3 label to enable
      community-based sepsis detection research\footnote{See
      \url{https://github.com/BorgwardtLab/mgp-tcn}
for more details.}.
    \item We present a detailed experimental setup in which we empirically demonstrate that our methods
    outperform the state of the art in detecting sepsis \emph{early}.
\end{itemize}

\paragraph{Technical Significance}
Ours is the first work to combine the MGP adapter framework~\citep{bonilla2008multi, li2016scalable}
with TCNs~\citep{lea2017temporal}, thus improving memory efficiency,
scalability, and classification performance for sepsis early detection
on irregularly-sampled time series.
We outperform the state-of-the-art method and improve AUPRC from $0.25$ to $0.35$/$0.40$, respectively~(measured \SI{7}{\hour} before sepsis onset).

\paragraph{Clinical Relevance}
The delayed identification and treatment of sepsis is a major driver of mortality in the ICU. By detecting sepsis \emph{earlier}, our approach could significantly decrease
mortality, because timely management is essential in this context~\citep{ferrer2014empiric}. An early warning system that is based on our methods could prevent delays in the initiation of antimicrobial and supportive therapy, which are considered to be crucial for improving patient outcome~\citep{kumar2006duration}. 

\section{Related Work}

This section introduces the recent literature and current challenges for
sepsis detection.

\subsection{Supervised Learning on Medical Time Series}%

Supervised learning on time series datasets has been haunted by the crux
that labels per time point are often missing, especially in medical
applications \citep{reddy2015healthcare}. This hindrance also applies
to the early detection of sepsis. In previous work, it was usually
circumvented by applying ad-hoc schemes to determine resolved sepsis
labels \citep{calvert2016, mao2018multicentre, kam2017learning}. These
papers used a global time series label, such as an ICD disease code
intended for billing, and they estimated sepsis onset with easily
computable ad-hoc criteria.
However, when using such a patchwork label, it is unclear if
the patient \emph{actually} suffered from an event at this time and
not, for instance, one week later. 

By extracting Sepsis-3, which is the most recent sepsis criterion
\citep{singer2016third} that allows for temporal resolution,
we contribute a solution to this issue that continues to affect the study of machine learning for healthcare.
Even though some datasets~\citep{futoma2017learning,
desautels2016prediction} have high-resolution sepsis labels, they are
currently \emph{not} accessible to the research community. This leads to
reproducibility and comparability issues.
Thus, there are massive hurdles to overcome before novel approaches for
sepsis detection can be developed and thoroughly validated.

\subsection{Algorithms for the Early Detection of Sepsis}\label{sec:SEDA}

\paragraph{Overview}
In the last decade, several data-driven approaches for detecting sepsis
in the ICU have been presented \citep{desautels2016prediction,
calvert2016, kam2017learning, futoma2017learning, shashikumar2017}.
Many approaches selectively compare with simple clinical scores, such as
SIRS, NEWS  or MEWS~\citep{bone1992definitions, williams2012national,
stenhouse2000prospective}. However, none of these scores are intended as specific,
continuously-evaluated risk scores for sepsis.
Specifically, the SIRS criteria are now considered by clinicians to be unspecific and obsolete for the definition of sepsis~\citep{beesley2015we, kaukonen2015systemic}.
As an alternative to these scores,
\citet{henry2015} introduced a targeted real-time warning
score~(\mbox{TREWScore}) to predict septic shock, which is a frequent
complication following from sepsis. 
Notably, while many machine learning methods have surpassed generic or
simplistic clinical schemes, next to no papers actually performed the
hard comparison to other machine learning approaches in the literature.
As an exception, the application of LSTMs~\citep{kam2017learning} have been shown to be an improvement over the InSight model~\citep{calvert2016}, which is a regression
model with hand-crafted features. However, only reported
metrics were compared, whereas potentially differing processing pipelines
and label implementations~(which are closed source) could make a direct comparison problematic. These circumstances prompted us to baseline our work against state-of-the-art machine learning methods and on exactly the same sepsis early recognition pipeline, which we make publicly available.

\paragraph{State of the Art}

Sepsis detection methods are usually developed on real-world datasets
with prevalence values ranging from 6.6\% \citep{kam2017learning} to
21.4\% \citep{futoma2017learning}.
Despite this considerable class imbalance, to our knowledge, only
\citet{futoma2017learning} and \citet{desautels2016prediction} report the
area under the precision--recall curve~(AUPRC), in addition to the area
under the receiver operating characteristic curve~(AUC).
Given the class imbalance, AUC is known to be a less informative evaluation criterion \citep{saito2015precision}.
Thus, in terms of AUPRC, \citet{futoma2017learning} currently represent
the state of the art in the early detection of sepsis.
In a follow-up paper, \citet{futoma2017improved} improved their
performance by proposing task-specific tweaks, such as
label-propagation, additional feature extraction~(e.g.\ missingness
indicators), and separate task correlation matrices for their Gaussian Process. However, these extensions pertain to the input features and they modify the GP adapter framework as wrapped around their classifier, so they are orthogonal to our undertaking of improving the \emph{classifier} inside the GP adapter framework.

\subsection{Gaussian Process Adapters}

\citet{li2016scalable} showed that optimizing a Gaussian Process
im\-put\-ation of a time series end-to-end using the gradients of
a subsequent classifier leads to better performance than optimizing both
the classifier and the GP separately. This method, which is also referred to as
GP adapters, is not restricted to imputing missing data~\citep{li2017targeting}.
Recently, \citet{futoma2017learning} demonstrated that GP adapters are
a well-suited framework to handle the irregularly spaced time series in
early sepsis detection. Specifically, they confirmed earlier findings
\citep{li2016scalable} that in time series classification, GP adapters
outperform conventional GP imputation schemes that require a separate
optimization step, which is not driven by the classification task.

\section{Methods}

In the following, we describe our proposed MGP-TCN and DTW-KNN
methods\footnote{Our notation uses regular font for scalars, bold
lower-case for vectors, and bold upper-case for matrices.}. First,
Section~\ref{Sec:Proposed} gives a high-level overview of our deep
learning method\footnote{Please refer to Supplementary
Section~\ref{Sec:MGPAdapters} for more details on the end-to-end MGP adapter
framework.}, emphasizing the MGP component (i.e.,\ the first building
block of the method) which as a whole was previously applied
by~\citet{futoma2017learning}.
Section~\ref{Sec:TCN} then describes temporal
convolutional networks~(TCNs), the second building block.  %
Finally, Section \ref{sec:DTW-KNN} describes DTW-KNN. 

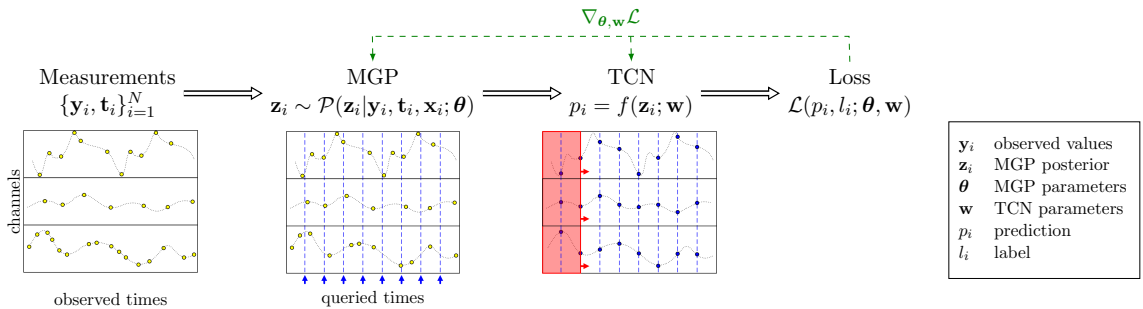
\begin{figure}[tbp]
\centering
\begin{adjustbox}{width=\textwidth}
\begin{tikzpicture}[
every matrix/.style={ampersand replacement=\&,column sep=1.4cm,row sep=2cm},
source/.style={draw,thick,rounded corners,fill=yellow!20,inner sep=.3cm},
process/.style={draw,thick,circle,fill=blue!20},
sink/.style={source,fill=red!5},
datastore/.style={draw,very thick,shape=datastore,inner sep=.3cm},
dots/.style={gray,scale=2},
to/.style={->,>=stealth',shorten >=5pt,ultra thick,font=\sffamily\footnotesize},
every node/.style={align=center}]

\matrix{
	\node[rectangle] (measurem) {Measurements \\ $\{ \mathbf{y}_i, \mathbf{t}_i \}_{i=1}^N $};
	\& \node[rectangle] (mgp) {MGP   \\ 
									$\mathbf{z}_i \sim \mathcal{P}(\mathbf{z}_i | \mathbf{y}_i, \mathbf{t}_i, \mathbf{x}_i;  \bm{\theta}  )$
		
									};	\&
    \node[rectangle] (tcn) {TCN \\ 
	    $p_i = f(\mathbf{z}_i; \mathbf{w}) $
									};  \&
	\node[rectangle] (loss) {Loss \\
		$\mathcal{L}(p_i, l_i ; \bm{\theta}, \mathbf{w}) $ 	
	};	
									\\
};

\draw[vecArrow] (measurem) -- node[midway,right] {}
node[midway,right] {} (mgp);
\draw[vecArrow] (mgp) -- node[midway,right] {} (tcn);
\draw[vecArrow] (tcn) -- node[midway,right] {} (loss);

\draw [black!50!green, dashed] (loss.north) -- (6.55,1) -- (-1.81,1) node[midway, above]{\small{$\nabla_{\bm{\theta},\mathbf{w}} \mathcal{L}$}};
\draw [black!50!green, dashed, -latex] (-1.81,1) -- (mgp.north);
\draw [black!50!green, dashed, -latex] (2.72,1) -- (tcn.north);

\node[inner sep=0pt] (ts1) at (-6.4,-2.0)
{\includegraphics[width=.2\textwidth]{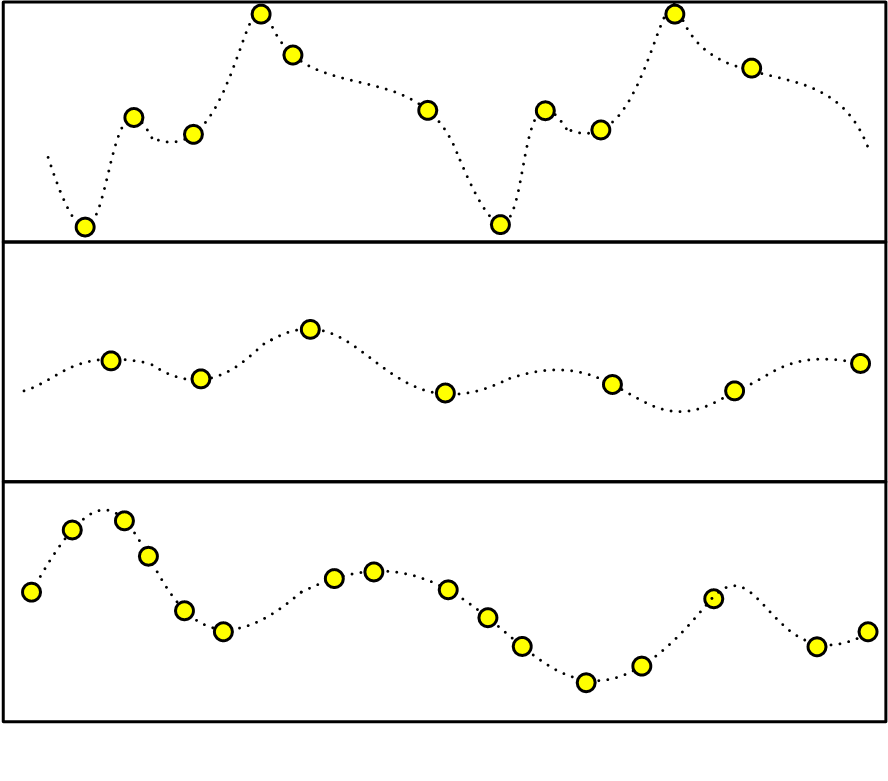}};
\node[scale=0.8, rotate=90] at (-8.1,-1.9) {channels};
\node[scale=0.8] at (-6.4,-3.6) {observed times};
\node[inner sep=0pt] (ts2) at (-1.80,-2.0)
{\includegraphics[width=.2\textwidth]{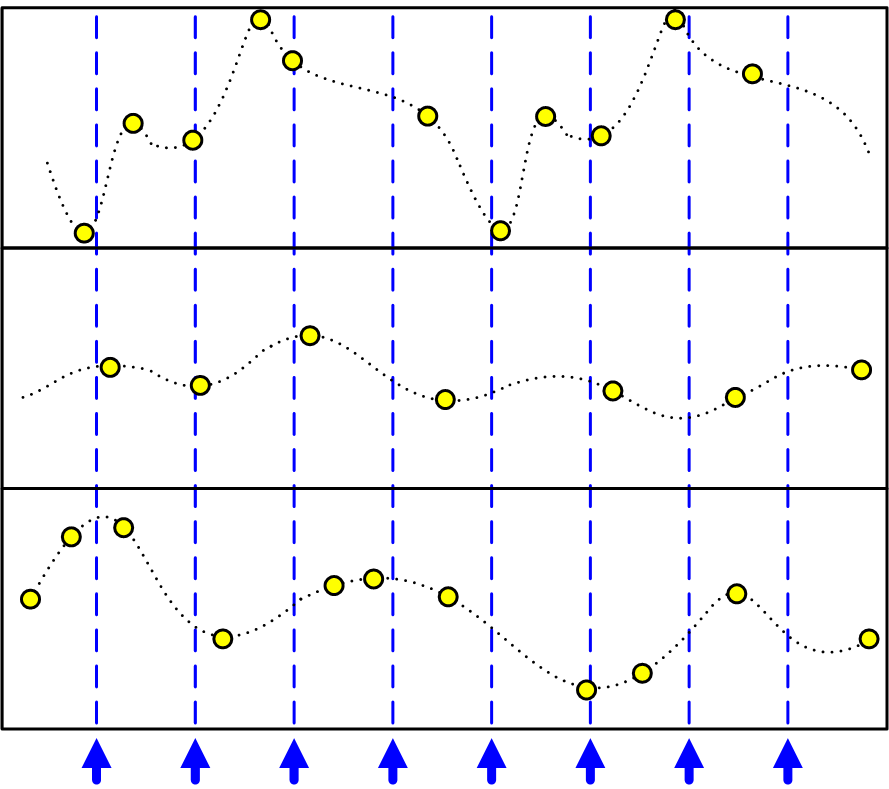}};
\node[scale=0.8] at (-1.80,-3.6) {queried times};

\node[inner sep=0pt] (ts3) at (2.7,-2.0)
{\includegraphics[width=.2\textwidth]{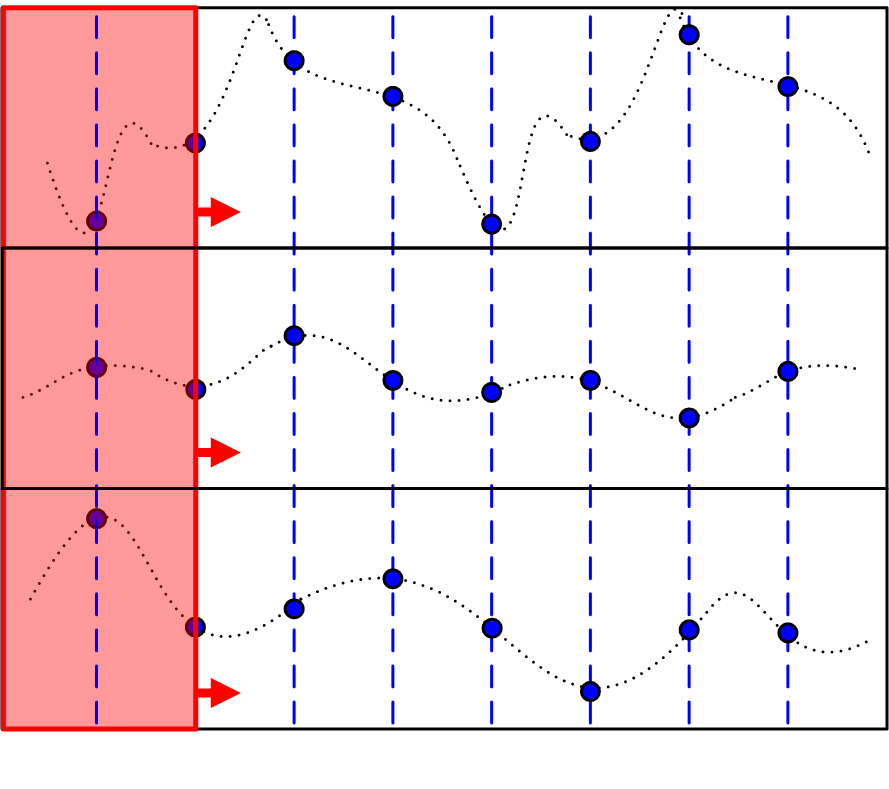}};

\draw [line width=0.02mm] (8.30,-3.25) rectangle (11.65, -0.5);
\node [scale=0.8, align=left] (variables) at (10.0, -1.85) {
																	\begin{tabular}{ll}
																	 $\mathbf{y}_i$ & observed values \\
																	 $\mathbf{z}_i$ & MGP posterior \\
																	 $\bm{\theta}$ & MGP parameters \\
																	 $\mathbf{w}$ & TCN parameters \\ 
																	 $p_i$ & prediction \\ 
																	 $l_i$ & label \\ 
																	 \end{tabular}
																	};

\end{tikzpicture}
\end{adjustbox}
\vspace*{-0.6cm}
\caption{%
  Overview of our model. Raw, irregularly-spaced time series are
  provided to the multi-task Gaussian Process~(MGP) for each patient.
  The MGP then draws from a posterior distribution, given the observed
  data, at evenly-spaced grid times~(each hour).
  This grid is then fed into a temporal convolutional network~(TCN)
  which, after a forward pass, returns a loss. Its gradient is then
  computed by backpropagation through both the TCN and the MGP~(green
  arrows).
  All parameters are learned end-to-end during training.
}
\label{Fig:Overview}
\end{figure}

\subsection{Multi-task Gaussian Process Temporal Convolutional Network Classifier} \label{Sec:Proposed}

We frame the early detection of sepsis in the ICU as a multivariate time series
classification task. Specifically, we focus on the task of identifying
sepsis onset in irregularly-sampled multivariate time series of
physiological measurements in ICU patients. Our proposed model uses
a multi-task Gaussian Process~(MGP) \citep{bonilla2008multi} that is
intrinsically capable of dealing with non-uniform sampling frequencies.
In this setting, the tasks considered by the MGP are the individual
channels of the time series.
More precisely, given irregularly-observed measurements (values and times) $\{ \mathbf{y}_i,
\mathbf{t}_i \}$ of encounter $i$, for evenly-spaced query times $\mathbf{x}_i$, the MGP draws a latent time series
$\mathbf{z}_i $ following the MGP's posterior distribution
$\mathcal{P}(\mathbf{z}_i | \mathbf{y}_i, \mathbf{t}_i, \mathbf{x}_i;
\bm{\theta}) $~(see Equation~\ref{eq:MGP overview}).
$ \mathbf{z}_i$ then serves as the input to
a temporal convolutional network~(TCN, Section~\ref{Sec:TCN}) that
predicts the sepsis label. Making use of the Gaussian Process adapter
framework~\citep{li2016scalable} enables us to optimize this entire process
end-to-end with respect to the final classification objective; that is,
identifying sepsis.
Figure~\ref{Fig:Overview} gives a high-level overview of the MGP-TCN model.
The MGP's posterior distribution follows a multivariate normal
distribution, i.e.\
\begin{equation}
    \mathbf{z}_i \sim \mathcal{N}\big( \bm{\mu}(\mathbf{z}_i), \mathbf{\Sigma}(\mathbf{z}_i); \bm{\theta} \big),
    \label{eq:MGP overview}
\end{equation}
with mean and covariance
\begin{align}
    \bm{\mu}(\mathbf{z}_i) &= ( \mathbf{K}^D \otimes \mathbf{K}^{X_i T_i}) ( \mathbf{K}^D \otimes \mathbf{K}^{T_i} + \mathbf{D} \otimes \mathbf{I} )^{-1} \mathbf{y}_i \\
    \begin{split}
    \mathbf{\Sigma}(\mathbf{z}_i) &= (\mathbf{K}^D \otimes \mathbf{K}^{X_i}) - (\mathbf{K}^D \otimes \mathbf{K}^{X_i T_i}) (\mathbf{K}^D \otimes \mathbf{K}^{T_i} + \mathbf{D} \otimes \mathbf{I})^{-1} (\mathbf{K}^D \otimes \mathbf{K}^{T_i X_i}).
    \end{split}
\end{align}
Here, $ \mathbf{K}^{X_i T_i} $ refers to the correlation matrix between
the evenly-spaced query times $\mathbf{x}_i$ and the observed times
$\mathbf{t}_i$, while $\mathbf{K}^{X_i}$ represents the correlations
between $\mathbf{x}_i$ with itself. $\mathbf{K}^D$ is the
task-similarity kernel matrix whose entry $K_{d,d'}^D$ at position $(d,
d')$ represents the similarity of tasks~(i.e.,\ time series channels) $d$ and $d'$. $\otimes$ denotes
the Kronecker product, and $\mathbf{K}^{T_i}$ represents an
encounter-specific $T_i \times T_i$ correlation matrix between all
observed times $t_i \in \mathbf{t}_i$ of patient encounter $i$, while
$\mathbf{D}$ is a diagonal matrix of per-task noise variances satisfying
$D_{dd} = \sigma_d^{2}$ and $\mathbf{I}$ refers to the identity matrix. The posterior mean $\bm{\mu}(\mathbf{z}_i)$ also depends on the observed values $\mathbf{y}_i$. We gather the MGP's parameters in $\bm{\theta} = \{ \mathbf{K}^D,{\sigma_d^{2}}|_{d=1}^{D},l  \}$ where $l$ refers to the length scale of the kernel function. For more details, please refer to Section \ref{Sec:MGPAdapters}.

\subsection{Temporal Convolutional Networks}\label{Sec:TCN}

\begin{figure}[tbp]
\centering
\includegraphics[width=0.7\linewidth]{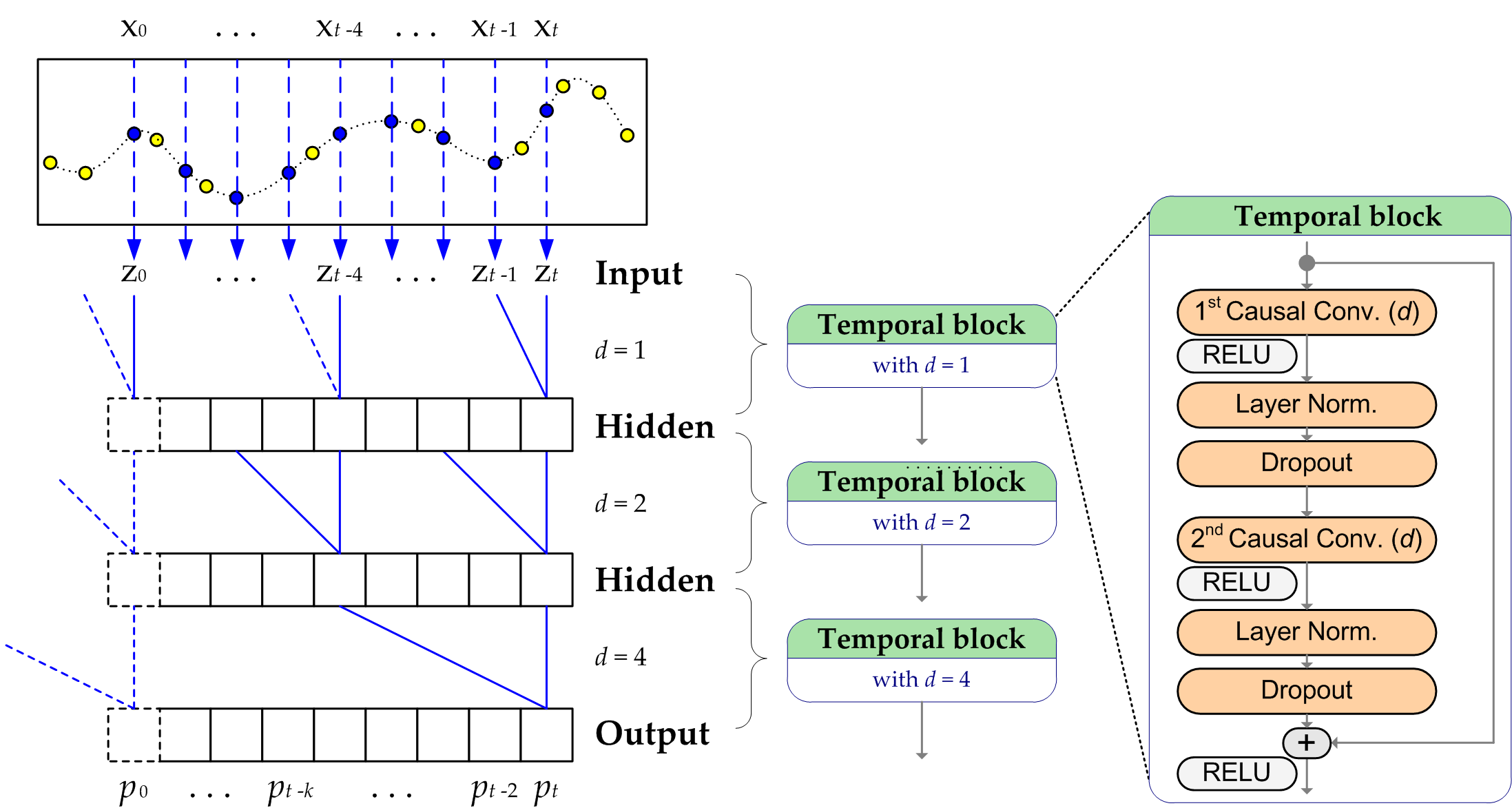}
\caption{%
Schematic illustration of the TCN architecture. The input $\mathbf{z}_i$ values~(blue) of the TCN classifier are computed by the multi-task Gaussian Process on a regular grid $( x_0,
\dots, x_t )$ based on the observed values~(yellow). Each temporal block skips an
increasing number of the previous layer's outputs, such that the \emph{visible
window} of a single node increases exponentially with increasing number of
layers. Figure recreated from \citet{bai2018empirical}.
}
\label{fig:tcn-architecture}
\end{figure}

This section outlines the details of a generic temporal convolutional
network~(TCN) architecture.
TCNs have recently been proposed~\citep{lea2017temporal} as an extension
of convolutional neural networks~(CNNs), which are known to exhibit state-of-the-art
performance in visual tasks~\citep{ciresan2011flexible, cirecsan2012multi}.
An empirical study by \citet{bai2018empirical} demonstrated that TCNs
show superior performance for sequence modeling tasks, as compared to
recurrent neural networks.
Please see Figure~\ref{fig:tcn-architecture} for an illustration of our
TCN architecture, for which the subsequent sections provide more
details.

\paragraph{Causal Dilated Convolutions}
TCNs are a simple but powerful extension to conventional
1D-CNNs in that they exhibit three properties~\citep{bai2018empirical}:
\begin{enumerate}
    \item \textbf{Sequence to sequence:} The output of a TCN has the same length as its input.
    \item \textbf{Convolutions are causal:} Outputs are only influenced
    by present and past inputs.
    \item \textbf{Long effective memory:} By using dilated convolutions,
    the receptive field, and thus also the effective memory, grows
    exponentially with increasing network depth.
\end{enumerate}
For 1.\ and 2., we use zero-padding to enforce equal layer sizes
throughout all layers while ensuring that for the output at time $t$,
only input values at this time and earlier times can be used~(see Figure~\ref{fig:tcn-architecture}). For 3., we follow the approach of
\citet{yu2015multi} by defining the $l$-dilated convolution of an input
sequence $\mathbf{x}$ with a filter $\mathbf{f}$ as
\begin{equation}
    \left( \mathbf{x}~*_{l}~\mathbf{f} \right)(k) = \sum_{k=i+l \cdot j} x_i \cdot f_j,
\end{equation}
where the $1$-dilated convolution coincides with the regular
convolution. By stacking $l$-dilated convolutions in an exponential
manner, such that $l = 2^n$ for the $n$-th layer, a long effective
memory can be achieved, as illustrated by~\citet{bai2018empirical}.

\paragraph{Residual Temporal Blocks}
We stack convolutional layers of a TCN into residual temporal blocks;
that is, blocks that combine the previous input and the result of the
respective convolution with an addition. Thus, the output of a temporal
block is computed relatively with respect to an input.
Here, we follow the setup of \citet{ceshine2018}, which applies layer
normalization~\citep{ba2016layer} to improve training stability and convergence,
as opposed to the weight normalization employed by \citet{bai2018empirical}.
Furthermore, we apply normalization \emph{after} each activation, similar to
\citet{ceshine2018}.

\subsection{Dynamic Time Warping Classifier}
\label{sec:DTW-KNN}

Dynamic time warping \citep{Keogh99} for time series classification,
using $k$-nearest neighbor approaches (here referred to as DTW-KNN), is
known to exhibit highly competitive predictive
performance \citep{dau2017judicious, ding2008querying}.
As opposed to many other off-the-shelf classifiers, it can handle
variable-length time series. Despite
its wide-spread use and demonstrated capabilities in data mining,
DTW-KNN has~(to the best of our knowledge) never been used in sepsis detection tasks.
We thus extend DTW-KNN for the classification of multivariate time
series, thereby introducing an additional novel approach for the early detection of sepsis. More precisely, we address the multivariate
nature of our setup by computing the DTW distance matrix~(containing the
pairwise distances between all patients) for each time series channel
separately. Each distance matrix is subsequently used for training a
$k$-nearest neighbor classifier.
Instead of using resampling techniques, the ensemble is constructed by
combining \emph{all} per-channel classifiers. Finally, for the classification
step, the final prediction score is computed as the average over all
per-channel prediction scores.

\section{Experiments}

\begin{figure}[tbp]
\begin{center}
\includegraphics[width=0.6\linewidth]{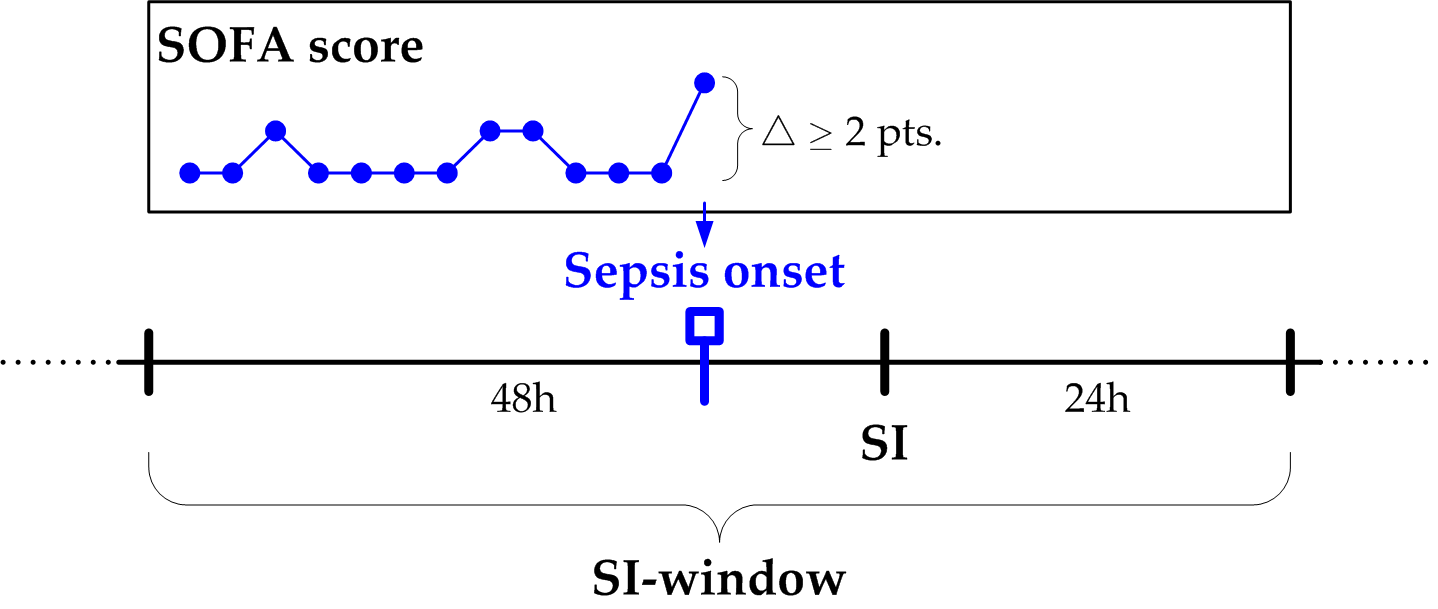}
\end{center}
\vspace*{-0.7cm}
\caption{
  For each encounter with a suspicion of infection~(SI), we extract a \SI{72}{\hour} window around the first SI event~(starting \SI{48}{\hour} before) as the SI-window.
  The Sequential Organ Failure Assessment (SOFA) score is then evaluated for every hour in this window by combining physiological scores of six organ systems. Following the SOFA definition, to arrive at a SOFA score we considered the worst organ scores of the last \SI{24}{\hour}.
}
\label{fig:sepsis-3}
\end{figure}

\subsection{Dataset and Sepsis Label Definition}

Our analysis uses the MIMIC-III~(\emph{Multiparameter Intelligent
Monitoring in Intensive Care}) database, version
1.4~\citep{johnson2016mimic}. MIMIC-III includes over 58,000 hospital
admissions of over 45,000 patients, as encountered between June 2001
and October 2012. 
We follow the most recent sepsis definition \citep{singer2016third},
which requires a co-occurrence of suspected infection~(SI) and organ
dysfunction. For SI, we follow the recommendations of
\citet{seymour2016assessment} to implement the SI cohort~(please
refer to Supplementary Section~\ref{Supp:Suspofinf} for more details).

According to \citet{singer2016third}, the organ dysfunction criterion is
fulfilled when the SOFA Score
\citep{vincent1996sofa} shows an increase of at least $2$ points.
To determine this, we follow the suggestions of
\citeauthor{singer2016third} to use
a window of \SIrange{-48}{24}{\hour} around a suspicion
of infection.
Figure~\ref{fig:sepsis-3} illustrates our Sepsis-3 implementation.
To detect sepsis \emph{early}, determining the sepsis onset
time is crucial.
We thus considerably refined and extended the queries provided by
\citet{Johnson18} to determine the Sepsis-3 label on an hourly basis\footnote{%
Their provided code only checks whether a simplified version of Sepsis-3
is satisfied upon admission. For instance, no \emph{increase} in SOFA
points is considered, but only one abnormally high value.}.
If sepsis is determined by merely checking whether a patient fulfills
the criteria upon admission, similarly to how it is done by
\citet{Johnson18}, then only those patients who \emph{arrive} in the ICU with
sepsis would be defined as cases, not the---arguably more interesting
ones---that \emph{develop} the syndrome during their ICU stay.

\subsection{Data Filtering}

\begin{table}[h] 
	\caption{%
    Characteristics of the population included in the dataset. The mean
    sepsis onset is given in hours since admission to the ICU.
  }
  \centering
	\begin{tabular}{lcc}
		\toprule
		\textbf{Variable} & \textbf{Sepsis Cases} & \textbf{Controls} \\
    \midrule
		n 	   &	570	      & 5,618  \\
		Female & 236 (41.4\%) & 2,548 (45.4\%) \\
		Male & 334 (58.6\%) & 3,070 (54.6\%) \\
    \midrule
    Mean time to sepsis onset in ICU (median) & \SI{16.7}{\hour} (\SI{11.8}{\hour}) & {---} \\
		Age ($\mu \pm \sigma$) & 67.2  $\pm$ 15.3 &  64.2 $\pm $  17.3 \\
    \midrule
    \textbf{Ethnicity} & & \\
    \midrule
        White                           &  411 (72.1\%) &  4,047 (72.0\%)\phantom{,} \\
        Black or African-American       &   41 (7.2\%) &   551 (9.8\%) \\
        Hispanic or Latino              &  \hspace{0.50em}7 (1.2\%) &   147 (2.6\%) \\
        Other                           &   \hspace{0.50em}57 (10.0\%) &   493 (8.8\%) \\
        Not available                   &   54 (9.5\%) &   380 (6.8\%) \\
		\midrule
    \textbf{Admission type} & & \\
    \midrule
		Emergency  &   504 (88.4\%) & 4,689 (83.5\%) \\
    Elective   &   \hspace{0.5em}60 (10.5\%) &  \hspace{0.60em}872 (15.5\%) \\
    Urgent     &   \hspace{0.50em}6 (1.1\%) &   \hspace{0.60em}57 (1.0\%) \\
		\bottomrule
	\end{tabular}
	\label{tab:Stats}
\end{table}

\paragraph{Patient Inclusion Criteria}
We exclude patients under the age of 15 and those for which no chart
data is available---including ICU admission or discharge time.
Furthermore, following the recent sepsis literature, ICU encounters
logged via the CareVue system were excluded due to underreported
negative lab measurements \citep{desautels2016prediction}.
We include an encounter as a case if at any time during the ICU stay
a sepsis onset occurs, whereas controls are defined as those patients
that have no sepsis onset~(they still might have suspected infection or
organ dysfunction, separately). To additionally ensure that controls
cannot be sepsis patients that developed sepsis shortly before ICU, we
require controls not to be labeled with any sepsis-related ICD-9 billing
code. Following these inclusion criteria, we initially count 1,797
sepsis cases and 17,276 controls.
This works aims for sepsis \emph{early} detection, so we follow
\citet{desautels2016prediction} and exclude cases that develop sepsis
earlier than seven hours into the ICU stay. This enables a prediction
horizon of \SI{7}{\hour}. To preserve a realistic class balance of around
10\%, we apply this exclusion step only after the case--control
matching~(see next paragraph). Thus, after cleaning and filtering, we finally use
570 cases and 5,618 controls as our cohort; Table~\ref{tab:Stats}
shows the summary statistics.
For the variables, we used 44 irregularly-sampled laboratory and vital
parameters\footnote{Please see Supplementary Section~\ref{Supp:vars} for
more details.}.
Furthermore, to be able to run all baselines, we had to apply an
additional patient filtering step\footnote{Please see Supplementary
Section~\ref{Supp:HorizonAdd} for more details.}.

\paragraph{Case--control Matching}
In previous work, it has been observed that an insufficient alignment of
time series of sepsis cases versus controls could render the
classification task trivial: for instance, when comparing a window
before sepsis onset to the last window~(before discharge) of a
control's ICU stay, the classification task is much easier than when
compared to a more similar reference time in a control's stay. This
can be observed by the decrease in performance of the MGP-RNN method
when \citet{futoma2017improved} applied case--control matching.
Hence, to avoid a trivial classification task, we also use
a case--control alignment in a matching procedure. To accommodate the
class imbalance, we assign each case to 10 random unassigned controls
and define their control onset as the absolute time~(in hours since
admission) when the matched case fulfilled the sepsis criteria.
\citet{futoma2017improved} used a relative measure; that is, the same
percentage of the entire ICU stay as for control onset time. However, we
observed that cases and controls do not necessarily share the same
length of stay, which could introduce bias to the alignment that a deep
neural network could potentially exploit. For each case and its matched controls, we extract up to \SI{48}{\hour} of input data preceding their onset and after ICU admission.

\subsection{Experimental Setup}
\label{Sec:exp_setup}

\paragraph{Baselines}
We compare our methods against MGP-RNN, which is the current state-of-the-art
sepsis detection method~\citep{futoma2017learning, futoma2017improved}.
To enable a fair comparison, the authors kindly provided source code for
their pipeline such that their model could be trained from scratch on
our dataset. Additionally, we compare against a classical TCN~(here referred
to as \mbox{Raw-TCN}) that is not embedded in the MGP adapter
framework.
To this end, as a preprocessing step, we first impute the times series
using a carry-forward scheme~(for more details, please refer to
Supplementary Section~\ref{Supp:Imput}). We train our DTW-KNN ensemble
classifier using the same imputation scheme.

\paragraph{Training}
We apply three iterations of random splitting using 80\% of the samples for
training and each 10\% for validation and testing. In each random split,
the time series were $z$-scored per channel using the respective train
mean and standard deviation.
For hyperparameter tuning, due to the costly evaluations, we apply an
off-the-shelf Bayesian optimization framework~(instead of an exhaustive
grid search) provided by the \texttt{scikit-optimize}
library~\citep{scikit-optimize} with 20 calls per method and split.
We select the best model parameters and checkpoints in terms of
validation AUPRC and we evaluate them on the test splits.
For all deep models, the hyperparameter spaces were constrained such
that the number of parameters each ranged from 20K--500K. To make it
feasible to analyze multiple random splits~(despite a deep learning
setup), we constrain each run to take at most two hours.
To prevent underfitting, we additionally retrained the best parameter
setting of each method and split for a longer period of 50
epochs.
Please refer to Supplementary Section~\ref{Supp:HyperSearch} for more
details.

\paragraph{Evaluation}
Due to substantial class imbalance~(the overall case prevalence is
9.2\%, or roughly 1 case for 10 controls), we evaluate all models in
terms of AUPRC on the test split. In addition, we report AUC, mostly
to comply with recent sepsis detection literature~(see also
Section~\ref{sec:SEDA} for a discussion of the disadvantages of this
measure).
Because timely identification of sepsis is of central importance, we
evaluate the trained models in a horizon analysis going back up to
\SI{7}{\hour} before sepsis onset. For example, to evaluate the
prediction horizon at \SI{3}{\hour} in advance, for each encounter, the
model (and imputation scheme) is only provided with input data up until that moment.
To assess the predictive performance, we do \emph{not} optimize the
models to each respective horizon hour, which would result in eight distinct
specialized models.
Instead, per method and fold, we train one model on all of the available
training data and challenge its performance by gradually restricting
access to the information closest to sepsis onset.

\paragraph{Implementation Details}
Additional information about the technical details of our implementation
and its runtime behavior are available in Supplementary
Section~\ref{supp:Implementation}.

\begin{figure}[t]

\definecolor{darkred}{rgb}{0.5450980392156862, 0.0, 0.0}
\definecolor{darkblue}{rgb}{0.0, 0.0, 0.5450980392156862}
\definecolor{darkgreen}{rgb}{0.0, 0.39215686274509803, 0.0}
\definecolor{violet}{rgb}{0.9333333333333333, 0.5098039215686274, 0.9333333333333333}
\centering
\begin{tikzpicture}
    \begin{groupplot}[%
    group style={group size=2 by 1, horizontal sep=1.5cm},
    x dir=reverse, 
    ymin=0, 
    ymax=0.7, 
    enlargelimits=false, 
    xlabel={Hours before sepsis onset}, 
    height=7cm, width=7cm, 
    legend pos=south east, 
    legend style = {draw = none, font = \tiny, }, 
    legend cell align = left, 
    try min ticks=10
    ] 
        \nextgroupplot[ylabel=AUPRC]
        %
        \addplot[no marks, darkred, very thick] table[col sep=comma, x=horizon, y=auprcs_mean]{data/MGP-TCN_results.csv};
        \addplot[no marks, darkblue, very thick] table[col sep=comma, x=horizon, y=auprcs_mean]{data/MGP-RNN_results.csv};
        \addplot[no marks, darkgreen, very thick] table[col sep=comma, x=horizon, y=auprcs_mean]{data/Raw-TCN_results.csv};
        \addplot[no marks, violet , very thick] table[col sep=comma, x=horizon, y=auprcs_mean]{data/DTW-KNN_results.csv};
        \legend{MGP-TCN, MGP-RNN, Raw-TCN, DTW-KNN}
        \addplot[no marks, darkred, draw opacity = 0.2, name path=upper] table[col sep=comma, x=horizon, y expr = \thisrow{auprcs_mean} + \thisrow{auprcs_std}]{data/MGP-TCN_results.csv};
        \addplot[no marks, darkred, draw opacity = 0.2, name path=lower] table[col sep=comma, x=horizon, y expr = \thisrow{auprcs_mean} - \thisrow{auprcs_std}]{data/MGP-TCN_results.csv};
        \addplot[darkred, fill opacity=0.1] fill between [of=upper and lower];
        %
        \addplot[no marks, darkblue, draw opacity = 0.2, name path=upper] table[col sep=comma, x=horizon, y expr = \thisrow{auprcs_mean} + \thisrow{auprcs_std}]{data/MGP-RNN_results.csv};
        \addplot[no marks, darkblue, draw opacity = 0.2, name path=lower] table[col sep=comma, x=horizon, y expr = \thisrow{auprcs_mean} - \thisrow{auprcs_std}]{data/MGP-RNN_results.csv};
        \addplot[darkblue, fill opacity=0.1] fill between [of=upper and lower];
        %
        \addplot[no marks, darkgreen, draw opacity = 0.2, name path=upper] table[col sep=comma, x=horizon, y expr = \thisrow{auprcs_mean} + \thisrow{auprcs_std}]{data/Raw-TCN_results.csv};
        \addplot[no marks, darkgreen, draw opacity = 0.2, name path=lower] table[col sep=comma, x=horizon, y expr = \thisrow{auprcs_mean} - \thisrow{auprcs_std}]{data/Raw-TCN_results.csv};
        \addplot[darkgreen, fill opacity=0.1] fill between [of=upper and lower];
        %
        \addplot[no marks, violet, draw opacity = 0.2, name path=upper] table[col sep=comma, x=horizon, y expr = \thisrow{auprcs_mean} + \thisrow{auprcs_std}]{data/DTW-KNN_results.csv};
        \addplot[no marks, violet, draw opacity = 0.2, name path=lower] table[col sep=comma, x=horizon, y expr = \thisrow{auprcs_mean} - \thisrow{auprcs_std}]{data/DTW-KNN_results.csv};
        \addplot[violet, fill opacity=0.1] fill between [of=upper and lower];
        
        \nextgroupplot[ymax=1.0, ylabel=AUC]
        \addplot[no marks, darkred, very thick] table[col sep=comma, x=horizon, y=aurocs_mean]{data/MGP-TCN_results.csv};
        \addplot[no marks, darkblue, very thick] table[col sep=comma, x=horizon, y=aurocs_mean]{data/MGP-RNN_results.csv};
        \addplot[no marks, darkgreen, very thick] table[col sep=comma, x=horizon, y=aurocs_mean]{data/Raw-TCN_results.csv};
        \addplot[no marks, violet , very thick] table[col sep=comma, x=horizon, y=aurocs_mean]{data/DTW-KNN_results.csv};
        \legend{MGP-TCN, MGP-RNN, Raw-TCN, DTW-KNN}
        \addplot[no marks, darkred, draw opacity = 0.2, name path=upper] table[col sep=comma, x=horizon, y expr = \thisrow{aurocs_mean} + \thisrow{aurocs_std}]{data/MGP-TCN_results.csv};
        \addplot[no marks, darkred, draw opacity = 0.2, name path=lower] table[col sep=comma, x=horizon, y expr = \thisrow{aurocs_mean} - \thisrow{aurocs_std}]{data/MGP-TCN_results.csv};
        \addplot[darkred, fill opacity=0.1] fill between [of=upper and lower];
        %
        \addplot[no marks, darkblue, draw opacity = 0.2, name path=upper] table[col sep=comma, x=horizon, y expr = \thisrow{aurocs_mean} + \thisrow{aurocs_std}]{data/MGP-RNN_results.csv};
        \addplot[no marks, darkblue, draw opacity = 0.2, name path=lower] table[col sep=comma, x=horizon, y expr = \thisrow{aurocs_mean} - \thisrow{aurocs_std}]{data/MGP-RNN_results.csv};
        \addplot[darkblue, fill opacity=0.1] fill between [of=upper and lower];
        %
        \addplot[no marks, darkgreen, draw opacity = 0.2, name path=upper] table[col sep=comma, x=horizon, y expr = \thisrow{aurocs_mean} + \thisrow{aurocs_std}]{data/Raw-TCN_results.csv};
        \addplot[no marks, darkgreen, draw opacity = 0.2, name path=lower] table[col sep=comma, x=horizon, y expr = \thisrow{aurocs_mean} - \thisrow{aurocs_std}]{data/Raw-TCN_results.csv};
        \addplot[darkgreen, fill opacity=0.1] fill between [of=upper and lower];
        %
        \addplot[no marks, violet, draw opacity = 0.2, name path=upper] table[col sep=comma, x=horizon, y expr = \thisrow{aurocs_mean} + \thisrow{aurocs_std}]{data/DTW-KNN_results.csv};
        \addplot[no marks, violet, draw opacity = 0.2, name path=lower] table[col sep=comma, x=horizon, y expr = \thisrow{aurocs_mean} - \thisrow{aurocs_std}]{data/DTW-KNN_results.csv};
        \addplot[violet, fill opacity=0.1] fill between [of=upper and lower];
        
    \end{groupplot}
\end{tikzpicture}
\caption{%
 We evaluate all methods using area under the precision--recall
 curve~(AUPRC) and additionally display the~(less informative) area
 under the receiver operating characteristic curve~(AUC).
 The current state-of-the-art method, MGP-RNN, is shown in blue. The two
 approaches for early detection of sepsis that were introduced in this
 paper (i.e.,\ MGP-TCN and DTW-KNN) are shown in pink and red,
 respectively. Using three random splits for all measures and
 methods, we show the mean~(line) and standard deviation error
 bars~(shaded area).
}
\label{fig:horizon}
\end{figure}
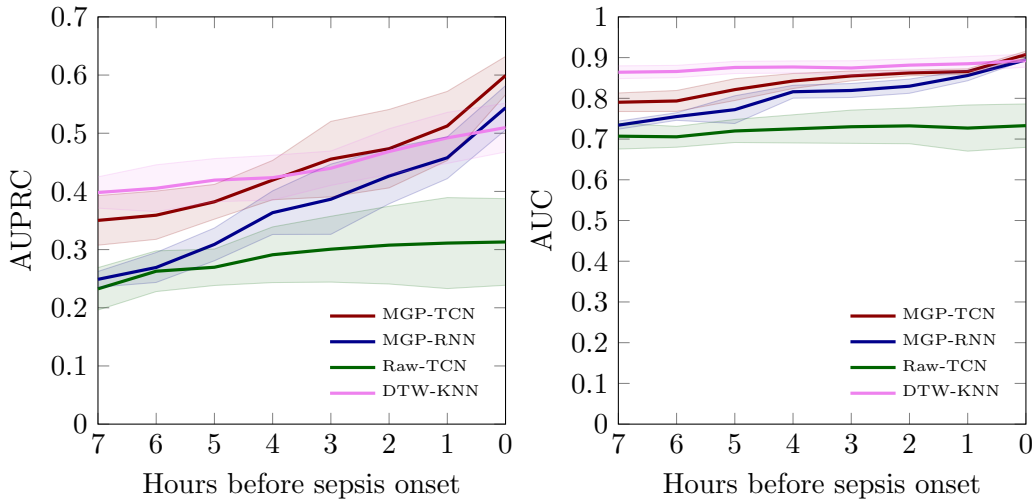

\subsection{Results}
Figure~\ref{fig:horizon} depicts the predictive performance for the
different time horizons.
The $x$-axes indicate the prediction horizon in hours before sepsis onset. The
$y$-axes measure AUPRC~(left) and AUC~(right). As previously discussed, we focus
on evaluating AUPRC due to the substantial class imbalance~(9.2\%).

We observe that both our novel model MGP-TCN and our DTW-KNN ensemble method
\emph{consistently} outperform the current state-of-the-art early sepsis
detection classifier MGP-RNN.
Especially for the early detection task of more than \SI{4}{\hour} before sepsis
onset, both MGP-TCN and DTW-KNN outperform the state of the art with
a significant margin, while the latter shows slightly higher performance in this
regime.
For horizons that are closer to sepsis onset, namely
\SIrange{0}{3}{\hour} prior to onset, \emph{all} approaches except
\mbox{Raw-TCN} exhibit overlapping performance in terms of their
variance estimates.
Interestingly, \mbox{Raw-TCN} does not yield competitive results for
\emph{any} setting that was considered in our experimental setup.  With
increasing distance to the onset, its performance starts to approximate
that of the MGP-RNN classifier. Finally, approaches based
on simple imputation schemes (i.e.,\ Raw-TCN and DTW-KNN) exhibit a much flatter
trend in AUPRC when approaching sepsis onset than the MGP-imputed ones.

\section{Conclusion}

Our proposed methods MGP-TCN and DTW-KNN exhibit favorable performances over all prediction horizons and they consistently outperform the
state-of-the-art baseline classifier MGP-RNN. Compared to the classic TCN, we empirically
demonstrated that TCN-based architectures---in combination with
MGPs to account for uncertainty associated with irregular sampling---result in competitive predictive performance.
Specifically, in terms of AUPRC, with MGP-TCN and DTW-KNN, we improve
the current state of the art from $0.25$ to $0.35$ and $0.40$,
respectively, \SI{7}{\hour} before sepsis onset.
This confirms that recent advances in sequence modeling may be
transferred successfully to irregularly-sampled medical time series.
By contrast, the low performance of the Raw-TCN classifier suggests that
a more advanced, uncertainty-aware imputation scheme is helpful when
transferring ``deep'' approaches to our scenario. Furthermore, we
observed that simple imputation schemes lead to a smaller gain in
performance when approaching sepsis onset. This could be due to the
nature of the carry-forward scheme, which tends to remove relevant
sampling information.

When comparing our findings with the recent literature, we observe that
the low prevalence in our dataset~(\SI{9.2}{\percent}) makes the classification task
substantially harder. For instance, in terms of AUPRC, MGP-RNN performed
better on its original dataset~(to which we unfortunately have no access),
which has a prevalence of \SI{21.4}{\percent}~\citep{futoma2017learning}.
In terms of prevalence and preprocessing, to our knowledge the most
comparable setup would be the one by \citet{desautels2016prediction};
however, we have made several requests to obtain their methods and queries, which have proven to be unsuccessful.
Interestingly, their reported AUPRC dropped to roughly $0.3$ already at
\SI{1}{\hour} before sepsis onset, where, for example, our proposed
MGP-TCN method still achieves an AUPRC of $0.51$.

One of the most surprising results is the highly-competitive performance
of our DTW-KNN ensemble classifier, whose performance for earlier horizons
exceeds all of the other methods, despite its conceptual simplicity.
While this is highly relevant for the early detection of sepsis, the DTW-KNN
classifier suffers from some practical limitations that impede online
monitoring scenarios and its application to very large patient cohorts. Already for our dataset, using a standard implementation, predicting at one horizon may
require hundreds of millions of pairs of univariate time
series to be aligned, followed by their distance computation, and the
subsequent storing of results~(which potentially affects both runtime
and memory). We conjecture that DTW-KNN performs so well because of the
mid-range sample size, whereas deep models tend to perform best for even
larger sample sizes.
However, scaling DTW-KNN to cohorts of hundreds of thousands of patients
currently appears to be a computational challenge. This also affects
online classification in the ICU: for each new measurement of a patient,
the distances of each involved channel to \emph{all} patients in the
training cohort have to be updated, and partially recomputed. Consequently, intermediate results of the distance calculations need to
be stored, leading to a significant memory overhead. The problem thus
remains a ``hot topic'' in time series analysis~\citep{oregi2017line}.

In contrast, MGP-TCN does not suffer from these limitations because only the
network weights have to be stored and classifying a new patient is
\emph{constant} in the total number of patients.  Thus, MGP-TCN can be
easily applied to larger cohorts, which is likely to further increase
predictive performance.  Moreover, obtaining online predictions only
requires very slight modifications. For more details on the scaling
behavior, please refer to Supplementary Section~\ref{Supp:scaling}.

\section{Future Work}

An additional source of validation of our findings would be to test our
model on more datasets. For the early detection of sepsis, this is
normally not done because the derivation of properly-resolved, time-based
sepsis labels requires considerable preprocessing efforts. For example,
in our work, the dynamic sepsis labels first required the implementation
of an entire query pipeline. Due to this bottleneck, the value of
providing publicly accessible sepsis labels for further research, as we
do in this work, cannot be overstated.
In the future, we also would like to extend our analysis to more types
of data sources arising from the ICU. \citet{futoma2017improved}
already employed a subset of baseline covariates, medication effects,
and missingness indicator variables.  However, a multitude of feature
classes still remain to be explored and integrated, each posing unique
challenges that will be interesting to overcome.
For instance, the combination of sequential and non-sequential features
has previously been handled by treating non-sequential features as 
sequential features~\citep{futoma2017learning}.
We hypothesize that this could be handled more efficiently by using
a more modular architecture that handles sequential and
non-sequential features differently.
Furthermore, we aim to obtain a better understanding of the time series
features used by the model. Specifically, we are interested in
assessing the \emph{interpretability} of the learned filters of the MGP-TCN
framework and then evaluate how much the activity of an individual filter
contributes to a prediction. This endeavor is somewhat facilitated by
our use of a convolutional architecture. The extraction of short
per-channel signals could prove very relevant for supporting diagnoses
made by clinical practitioners.

\raggedbottom

 \section*{Funding}

 This work was funded in part by the SPHN/PHRT Driver Project ``Personalized Swiss Sepsis Study'' as well as by the Alfried Krupp Prize for Young University Teachers of the Alfried Krupp von Bohlen und Halbach-Stiftung (K.B.).
\bibliography{main}


\clearpage
\appendix
\pagenumbering{Roman}

\counterwithin{figure}{section}
\counterwithin{table} {section}
\renewcommand\thesection{\Alph{section}}
\section{Supplementary Material}

\subsection{Multi-task Gaussian Process Adapters} \label{Sec:MGPAdapters}

\subsubsection{Multi-task Gaussian Processes}

We first describe how our approach models an individual time series with
potentially different sampling frequencies and missing observations.
To this end, we use a \emph{Gaussian Process}~(GP).
GPs are a popular choice to model time series because
they can handle variable spacing between observations. In addition, they
capture the predictive \emph{uncertainty} associated with missing data.
To account for multivariate time series, we make use of a MGP \citepappendix{bonilla2008multi} with the tasks
representing the different medical variables. 

Given a patient encounter $i$ fully-observed at $T_i$ times, we
``unroll'' the different channels of the time series, gathering the
values of $D$ variables in a vector \\ $\mathbf{y}_i = (y_{11}, \dots, y_{T_i1},
\dots, y_{12}, \dots, y_{T_i2}, \dots, y_{1D}, \dots, y_{T_iD})^T $, and
collect all $T_i$ observation times in a vector $\mathbf{t}_i$.  In
clinical practice, this array is sparse and hence inefficient to store
explicitly; we only use it here for notational convenience. Each
encounter receives a binary label $l_i$ indicating whether the patient
develops sepsis during this stay. \newline
We model the true value of encounter $i$ and variable $d$ at time $t$
using a latent function $f_{i,d}(t)$. The MGP places a Gaussian Process
prior over the latent functions to directly induce the correlation
between tasks using a shared correlation function $k^\tau(\cdot, \cdot)$
over time. Assuming zero-meaned GPs, we have:
\begin{align}
    \langle f_{i,d}(t), f_{i,d'}(t') \rangle &= K_{d,d'}^D~k^\tau(t,t') \\ 
    y_{i,d}(t) &\sim \mathcal{N}\left(f_{i,d}\left(t\right),\sigma_d^{2}\right),
\end{align}
where $\mathbf{K}^D$ is the task-similarity kernel matrix whose entry
$K_{d,d'}^D$ at position $(d, d')$ represents the similarity of tasks
$d$ and $d'$, while $\sigma_d^{2}$ denotes the noise variance of task
$d$. An entire, fully-observed multivariate time series of  encounter
$i$ follows
\begin{align}
    \mathbf{y}_i &\sim \mathcal{N}(\mathbf{0}, \mathbf{\Sigma}_i) \\ 
    \mathbf{\Sigma}_i &= \mathbf{K}^D \otimes \mathbf{K}^{T_i} + \mathbf{D} \otimes \mathbf{I},
\end{align}
where $\otimes$ denotes the Kronecker product, and $\mathbf{K}^{T_i}$
represents an encounter-specific $T_i \times T_i$ correlation matrix
between all observed times $t_i \in \mathbf{t}_i$ of encounter $i$,
while $\mathbf{D}$ is a diagonal matrix of per-task noise variances
satisfying $D_{dd} = \sigma_d^{2}$. Building on previous work on
modeling noisy physiological time series \citepappendix{williams2006gaussian,
futoma2017learning}, we use an Ornstein--Uhlenbeck kernel as
a correlation function; that is, $k^\tau(t,t'; l) := \exp(-|t - t'|/l)$,
parametrized using a length scale $l$. For simplicity, we share
$\mathbf{K}^D$ and $k^\tau(\cdot,\cdot; l)$ and the per-task noise
variances across different patients. Hence, the parameterization of
the MGP can be summarized as 
\begin{equation}
  \bm{\theta} = \{ \mathbf{K}^D,{\sigma_d^{2}}|_{d=1}^{D},l  \}
\end{equation}
or $D^2 + D + 1$ parameters. 
In a fully-observed setting, $\mathbf{\Sigma}_i$ is a $D \cdot T_i
\times D \cdot T_i$ covariance matrix. However, in clinical practice,
only a subset of all $D$ variables is measured at most observation
times. Thus, we only have to compute entries of  $\mathbf{\Sigma}_i$ for
\emph{observed} pairs of time and variable type. So for encounter $i$,
if the number of all observed measurements $m_i < D \cdot T_i$, we only
compute an $m_i \times m_i$ covariance matrix.

Following \citetappendix{futoma2017learning}, we use the MGP to preprocess the
sparse and irregularly spaced multi-channel time series of a patient's
measurements to output an regularly-spaced time series driven
by the final classification task. To achieve this, let $\mathcal{X}$ be
a list of regularly-spaced points in time that starts with the ICU
admission as hour zero and continues by counting the time since
admission~(in our case) in full hours. Using this grid, for each
encounter we derive a vector $\mathbf{x}_i = (x_{1}, \dots, x_{X_i} )$
of grid times which will be used as query points for the MGP. More
specifically, $x_1 = 0$ and $x_{n+1}-x_{n}=1 $ for all encounters. We
use the next full hour after the last observed point in time as the
encounter-specific last grid time $x_{X_i}$~(for more details on how we
select the patient time window, please refer to
Section~\ref{Sec:exp_setup}).
On a patient level, the MGP induces a posterior distribution over the $D
\times X_i $ matrix $\mathbf{Z}_i$ of imputed time series values at the
$X_i$ queried points in time for $D$ tasks. As previously
shown \citepappendix{bonilla2008multi, futoma2017learning}, when stacking the
columns of $\mathbf{Z}_i$ such that $\mathbf{z}_i
= \textrm{vec}(\mathbf{Z}_i)$, the posterior distribution follows
a multivariate normal distribution
\begin{align}
    \mathbf{z}_i &\sim \mathcal{N}\big( \bm{\mu}(\mathbf{z}_i), \mathbf{\Sigma}(\mathbf{z}_i); \bm{\theta} \big)\label{eq:MGP posterior}\\
    \shortintertext{with}
    \bm{\mu}(\mathbf{z}_i) &= ( \mathbf{K}^D \otimes \mathbf{K}^{X_i T_i})\mathbf{\Sigma}_i^{-1} \mathbf{y}_i \\
    \begin{split}
    \mathbf{\Sigma}(\mathbf{z}_i) &= (\mathbf{K}^D \otimes \mathbf{K}^{X_i}) \\ \ & - (\mathbf{K}^D \otimes \mathbf{K}^{X_i T_i})\mathbf{\Sigma}_i^{-1}(\mathbf{K}^D \otimes \mathbf{K}^{T_i X_i}).
    \end{split}
\end{align}
Here, $ \mathbf{K}^{X_i T_i} $ refers to the correlation matrix between
the queried grid times $\mathbf{x}_i$ and the observed times
$\mathbf{t}_i$ while $\mathbf{K}^{X_i}$ represents the correlations
between $\mathbf{x}_i$ with itself.

\subsubsection{Classification Task}

So far, we have outlined how the MGP returns an evenly-spaced
multi-channel time series $\mathbf{Z}_i$ when given a patient's raw time
series data $\{ \mathbf{y}_i, \mathbf{t}_i \} $.
To train a model and ultimately perform classification, we require a
loss function.
As \citetappendix{li2016scalable} stated first, if $\mathbf{Z}_i$ were directly
observed, it could be directly fed into a off-the-shelf classifier such
that its loss could be simply computed as $\mathcal{L}(f(\mathbf{Z}_i;\mathbf{w}), l_i)$
with $l_i $ denoting the class label.
However, $\mathbf{Z}_i$ is actually a random variable and so is the loss
function. We account for this by using the expectation $\E_{\mathbf{z}_i
\sim \mathcal{N}\left( \bm{\mu}(\mathbf{z}_i),
\mathbf{\Sigma}(\mathbf{z}_i); \bm{\theta}\right)
} [ \mathcal{L}(f(\mathbf{Z}_i;\mathbf{w}), l_i) ] $ as the overall loss
function for optimization. The learning task then becomes minimizing
this loss over the entire dataset. Thus, we search parameters
$\mathbf{w}^*, \bm{\theta}^*$ that satisfy:
\begin{align}
{
\mathbf{w}^*,\bm{\theta}^*= \argmin_{\mathbf{w},\bm{\theta}} \sum_{i=1}^N \overbrace{\rule{0pt}{0.5cm} \mathbb{E}_{\mathbf{z}_i \sim \mathcal{N}\left( \bm{\mu}(\mathbf{z}_i), \mathbf{\Sigma}(\mathbf{z}_i); \bm{\theta}\right) } \big[ \mathcal{L}(f(\mathbf{Z}_i;\mathbf{w}), l_i) \big] }^{\text{$E_i$}} 
}
\label{Eq:opt}
\end{align}
For many choices of $f(\cdot) $ the expectation $E_i$ in
Equation~\ref{Eq:opt} is analytically not tractable.
We thus use Monte Carlo sampling with $s$ samples to approximate this
term as
\begin{align}
E_i \approx \frac{1}{S} \sum_{s=1}^{S} \mathcal{L}(f(\mathbf{Z}_s;\mathbf{w}), l_i),
  \intertext{where}
  \textrm{vec}(\mathbf{Z}_s) = \mathbf{z}_s \sim \mathcal{N}\big( \bm{\mu}(\mathbf{z}_i), \mathbf{\Sigma}(\mathbf{z}_i); \bm{\theta} \big).
\end{align}
To compute the gradients of this expression with respect to both the
classifier parameters $\mathbf{w}$ and the MGP parameters $\bm{\theta}$,
we make use of a reparametrization \citepappendix{kingma2014} and set
$\mathbf{z}_i = \bm{\mu}(\mathbf{z}_i) + \mathbf{R}\bm{\xi} $, where
$\mathbf{R}$ satisfies $\mathbf{\Sigma}(\mathbf{z}_i)
=\mathbf{R}\mathbf{R}^T$ and $\bm{\xi} \sim
\mathcal{N}(\mathbf{0},\mathbf{I})$.
In this work, for the sake of simplicity, we use a Cholesky decomposition
to determine $\mathbf{R}$ and refrain from more involved approximative
techniques~(such as the Lanczos approach used by \citetappendix{li2016scalable}).

\subsection{Sepsis-3 Implementation}

Following \citetappendix{singer2016third} and \citetappendix{seymour2016assessment}, for
each encounter with a suspicion of infection (SI), for the first SI
event we extract the 72 hours window around it (starting 48 hours
before) as the SI-window.

\subsubsection{Suspicion of Infection}

To determine suspicion of infection, we follow
\citetappendix{seymour2016assessment}'s definition of the suspected infection
(SI) cohort. The SI criterion manifests in the timely co-occurrence of
antibiotic administration and body fluid sampling. If a culture sample
was obtained before the antibiotic, then the drug had to be ordered within 72
hours. If the antibiotic was administered first the sampling had to
follow within 24 hours. Here, we follow \citetappendix{Johnson18} to use the sampling to define the SI time, whereas \citetappendix{seymour2016assessment} indicated that the specific SI windowing is rather arbitrary and could be chosen differently. 

\subsubsection{Organ Dysfunction}

The SOFA score \citepappendix{vincent1996sofa} is a scoring system that is recommended by Sepsis-3 to assess organ dysfunction.  Given that this is
a particularly time-sensitive matter, we evaluate the SOFA score~(which
considers the worst parameters of the previous 24 hours) at each hour of
the 72 hour window around suspicion of infection. More importantly, as
Sepsis-3 foresees, to ensure an acute increase in SOFA of at least
two points, we trigger the organ dysfunction criterion when SOFA has
\emph{increased} by two points or more during this window.
\label{Supp:Suspofinf}

\subsection{List of Clinical Variables}
\label{Supp:vars}

Table~\ref{Supp:tableVars} lists all used clinical variables, comprising 44 vital and laboratory parameters.

\begin{table}[tbp]
  \caption{List of all 44 used clinical variables. For this study, we
  focused on irregularly sampled time series data comprising vital and
  laboratory parameters. To exclude variables that rarely occur, we selected variables with 500 or more observations present in the patients that fulfilled our original inclusion criteria (1,797 cases and 17,276 controls).
  }
	\label{Supp:tableVars}
	\centering
    \begin{tabular}{lll}
    \toprule
     \textbf{Vital Parameters} &  \\
    \midrule
     Systolic Blood Pressure & Tidal Volume Set  \\
     Diastolic Blood Pressure & Tidal Volume Observed  \\
     Mean Blood Pressure  & Tidal Volume Spontaneous \\
     Respiratory Rate & Peak Inspiratory Pressure  \\
     Heart Rate & Total Peep Level  \\
     SpO2 (Pulsoxymetry) &  O2 flow \\
     Temperature Celsius & FiO2 (Fraction of Inspired Oxygen) \\
     Cardiac Output & \\
     \midrule 
     \textbf{Laboratory Parameters} &  \\
     \midrule
     Albumin & Blood Urea Nitrogen  \\ 
     Bands (Immature Neutrophils) &  White Blood Cells \\
     Bicarbonate & Creatine Kinase  \\
     Bilirubin & Creatine Kinase MB  \\
     Creatinine & Fibrinogen \\
     Chloride & Lactate Dehydrogenase \\
     Sodium & Magnesium \\
     Potassium  &  Calcium (free) \\
     Lactate  & pO2 Bloodgas  \\
     Hematocrit &  pH Bloodgas \\
     Hemoglobin &   pCO2 Bloodgas  \\
     Platelet Count & SO2 Bloodgas  \\
     Partial Thromboplastin Time & Glucose  \\
     Prothrombin Time (Quick) &Troponin T \\  
     INR (Standardized Quick) &  \\

    \bottomrule

    \end{tabular}
    \vspace*{0.5cm}
\end{table}

\subsection{Imputation Schemes}
\label{Supp:Imput}

Here we provide more details about how the methods that do \emph{not} employ
an MGP were imputed. For maximal comparability to the MGP sampling
frequency, we binned the time series into bins of one hour width by
taking the mean of all measurements inside this window. We then apply
a carry-forward imputation scheme were empty bins are filled with the
value of the last non-empty one. The remaining empty bins~(at the start
of the time series) were then mean-imputed~(which after centering was
reduced to 0 imputation).

\subsection{Hyperparameter Search}

\paragraph{Differentiable models}
For the differentiable models \emph{MGP-RNN}, \emph{MGP-TCN}, and
\emph{Raw-TCN} an extensive hyperparameter search based on Bayesian
optimization was performed using the \texttt{scikit-optimize}
package \citepappendix{scikit-optimize}.
More precisely, we relied on a Gaussian Process to model
the AUPRC of the models dependent on the hyperparameters.
The models were then trained at the hyperparameter values that matched one of
the randomly-selected criteria \emph{largest confidence bounds},
\emph{largest expected improvement} and \emph{highest probability of
improvement} according to the Gaussian Process prior.
A total of ten initial evaluations were performed at random according to the
hyperparameter search space, followed by ten evaluations according to
the Gaussian Process prior. During the hyperparameter search phase, the
\emph{MGP-RNN} model was trained for
5 epochs over the complete dataset---since we observed fast convergence
behavior---while the TCN based model were trained 15 and 100 epochs
for the \emph{MGP-TCN} and the \emph{Raw-TCN} model respectively.

\begin{table*}[tbp]
\caption{Detailed information about hyperparameter search ranges. *: we fixed 10 Monte Carlo samples according to  \protect\citetappendix{futoma2017learning}. **: the MGP-RNN baseline was presented with a batch-size of 100, thus we did not enforce our range on this baseline \protect\citepappendix{futoma2017learning}.}
   \begin{tabular}{llccc}
        \toprule
        \textbf{Model} & \textbf{Hyperparameter} & \textbf{Lower bound} & \textbf{Upper bound} & \textbf{Sampling distribution} \\
        \midrule
        \multirow{2}{*}{All models}
            & learning rate & \num{5e-4} & \num{5e-3} & log uniform \\
            & Monte Carlo Samples &
                \multicolumn{2}{c}{\num{10}*} & not applicable\\
         \midrule
         \multirow{5}{*}{\shortstack{MGP-TCN\\ Raw-TCN}}
            & batch size & \num{10} & \num{40} & uniform \\
            & temporal blocks & 4 & 9 & uniform\\
            & filters per layer & 15 & 90 & uniform\\
            & filter width & 2 & 5 & uniform\\
            & dropout & 0 & 0.1 & uniform \\
            & $L_2$-regularization & 0.01 & 100 & log uniform \\
        \midrule
        \multirow{4}{*}{MGP-RNN}
            & batch size & \multicolumn{2}{c}{\num{100}**} & not applicable \\
            & layers & 1 & 3 & uniform \\
            & hidden units per layer & 20 & 150 & uniform \\
            & $L_2$-regularization & \num{1e-4} & \num{1e-3} & log uniform \\
        \bottomrule
    \end{tabular}

\end{table*}

\paragraph{DTW-KNN classifier}
The performance of the DTW-KNN classifier was evaluated on the same
validation dataset as the other models for $k \in \{1, 3, \dots, 13,
15\}$, while relying on the training dataset with 0 hours before Sepsis
onset. The $k$ value yielding the best AUPRC on the validation dataset
was selected for subsequent evaluation on the testing dataset. Similar
to the other classifiers, we do not ``refit'' the $k$-nearest neighbors
classifier by removing data from the training dataset to generate the
horizon plots.

\subsection{Additional Information on Scaling}

\label{Supp:scaling} \citetappendix{li2016scalable} demonstrated that the MGP adapter
framework is dominated by drawing from the MGP distribution. Thus, in our case, inverting
$\mathbf{\Sigma_i} \in \mathbb{R}^{D \cdot T_i \times D \cdot T_i}$ has
a computational complexity of $\mathcal{O}(D^3 \cdot T_i^3)$ \citepappendix{golub2012matrix}.
Notably, classifying a patient \emph{only} depends on the length of the
current patient time series and the number of tasks/variables; it does
\emph{not} depend on the number of patients in the training dataset; that is, it has a complexity of $\mathcal{O}(1)$ in the number of patients.

By contrast, while a naive implementation of DTW-KNN has a very low
training complexity~($\mathcal{O}(1)$, due to its character as a ``lazy
learner''), the complexity at prediction time is very high.
To classify a single instance,  the $k$-nearest neighbors
classifier requires the distances of the instance to all $N$ training
points.
Moreover, the runtime of a single distance computation using dynamic time warping~(DTW)
is $\mathcal{O}(D T^2)$, where $D$ represents the number of channels in the time
series and $T$ is the length of the shorter time series. Overall, this leads to
a runtime complexity of $\mathcal{O}(NDT^2)$ for a \emph{single}
prediction step, which can quickly become infeasible for large-sized
heath record datasets, especially if online predictions are desired.
Consequently, already for $N \geq D^2 T$, the complexity of the DTW-KNN approach
will exceed that of the MGP-TCN. Furthermore, the cubic complexity in
the prediction step can be ameliorated by using faster approximation
schemes \citepappendix{li2016scalable}.

\subsection{Supplementary Results}
\label{Supp:HyperSearch}

To enforce a maximum time of two hours per call for each method~(in order to make different hyperparameter searches on several folds feasible), MGP-RNN trains for 5 epochs, MGP-TCN for 15 epochs, and Raw-TCN for 100 epochs. We applied early stopping based on validation AUPRC with patience $=5$ epochs.

\begin{figure}[tbp]
\includegraphics[width=0.9\textwidth]{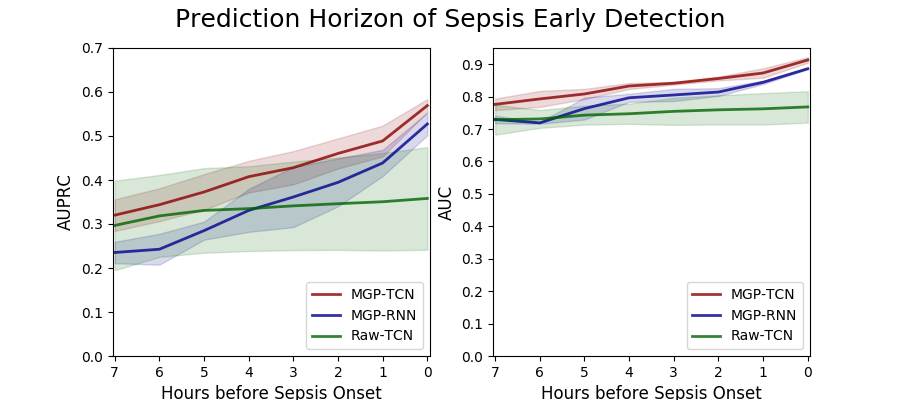}
\label{Supp:horizon}
\caption{
 The auxiliary setup as computed for the full horizon. We retrained the best parameter settings of all deep model for a prolonged time to guard against underfitting. Due to a slightly decreased performance, the MGP-based models show a tendency to overfit when training for 50 epochs, whereas the Raw-TCN shows high variance between the random splits.
}
\end{figure}

Moreover, in an auxiliary setup, to guard against underfitting, we use the best
parameter setting of each deep model and retrain each model for a prolonged
period~(50 epochs for both MGP-based models, 100 epochs for Raw-TCN)
using early stopping based on validation AUPRC with patience $=10$ epochs.
As shown in Figure~\ref{Supp:horizon}, the deep models exhibit a similar
tendency as in Figure~\ref{fig:horizon}, with the exception of the
Raw-TCN showing high variability~(due to one fold with favorable performance).
Also, the MGP-based models exhibit a slight drop in performance. This
may indicate that in terms of epochs, they converge earlier and overfit
earlier.

\subsection{Additional Information for the Horizon Analysis}
\label{Supp:HorizonAdd}

When creating the horizon analysis, we observed that the current MGP-RNN
implementation~(using a Lanczos iteration) requires the minimal number of
observed measurements of a patient to be at least the number of Monte Carlo
samples~(i.e.\ $10$ in our case). Hence, we performed an on-the-fly
masking of encounters that did \emph{not} satisfy this criterion; for
comparability, we applied it to all models.
Table~\ref{Supp:horizonStats} details the patient counts obtained
after masking. Additionally, to be able fit all methods into memory, we
removed a single outlier encounter consisting of more than 10K measurements.
Notably, because we did not refit the models on each horizon~(as this would
answer a different question), with increasing prediction horizon, the
slight decrease in sample size should not bias the mean performance but
rather scale up the error bars.

\begin{table}[H]
	\centering
	\caption{Patients count after applying masking which was required for making the MGP-RNN baseline work.}
	\begin{tabular}{cc ccc ccc ccc}
		\toprule
		\multirow{2}{*}{\textbf{Horizon}} & \textbf{Split} &
		\multicolumn{3}{c}{Train} & \multicolumn{3}{c}{Validation} &	\multicolumn{3}{c}{Test} \\
		& \textbf{Fold} & 0 & 1 & 2 & 0 & 1 & 2 & 0 &  1 & 2   \\
		\midrule
		0 && 4953 & 4950 & 4950 & 618 & 618 & 619 & 617 & 620 & 619 \\
		1 && 4951 & 4947 &4947 & 618 & 618 & 619 & 616 & 620 & 619 \\
		2 && 4943 & 4941 & 4937 & 616 & 617 & 619 & 616 & 617 & 619 \\
 		3 && 4933 & 4933 & 4927 & 614 & 615 & 617 & 615 & 614 & 618 \\
		4 && 4913 & 4917 & 4910 & 613 & 611 & 616 & 613 & 611 & 613 \\
		5 && 4832 & 4827 &  4830 & 602 & 605 & 602 & 598  & 600 & 600 \\
 		6 && 4587 & 4580 & 4565 & 566 & 570 & 581 & 567 & 570 & 574 \\
		7 && 4073 & 4061 & 4056 & 503 & 508 & 510 & 497 & 504 & 507 \\
	
			\bottomrule
	\end{tabular}
	\label{Supp:horizonStats}
\end{table}

\subsection{Implementation Details \& Runtimes}
\label{supp:Implementation}
For maximal reproducibility, we embedded our method and all comparison partners
in the \texttt{sacred v0.7.4}  environment \citepappendix{klaus_greff-proc-scipy-2017}.
A local installation of the MIMIC-III database was done with PostgreSQL 9.3.22.
The queries to extract the data from the database are based on queries in the
public \texttt{mimic-code} repository \citepappendix{Johnson18}.
However, to extract the hourly-resolved sepsis label, we had to implement
an entire query pipeline on top of the original code \citepappendix{Johnson18}.
For the MGP module, we included the code implemented by
\citepappendix{futoma2017learning} with minor changes.
For the TCNs, we extend the
TensorFlow implementation of \citetappendix{ceshine2018}.
We further apply \emph{gradient checkpointing} \citepappendix{chen2016training}
for all neural network models in order to permit training in a typical GPU setup.
The DTW-KNN classifier was implemented using the libraries
\texttt{tslearn v0.1.26} \citepappendix{tslearn} for dynamic time warping and
\texttt{scikit-learn v0.20.2} \citepappendix{scikit-learn} for the $k$-nearest
neighbor classifier.
We implemented both our proposed methods as well as all comparison
partners in Python 3. All experiments were performed on a Ubuntu 14.04.5
LTS server with 2 CPUs~(Intel\textsuperscript{\textregistered}
Xeon\textsuperscript{\textregistered} E5-2620 v4 @ 2.10GHz),
8 GPUs~(NVIDIA\textsuperscript{\textregistered} GeForce\textsuperscript{\textregistered} GTX 1080),
and 128 GiB of RAM.
However, for the deep learning models, we exclusively used single GPU
processing.
Supplementary Table~\ref{Supp:Runtimes} depicts the runtimes of all methods.

\begin{table}[H]
  \centering
  \caption{
    Training runtimes~(for the RNN/TCN methods, this includes the sum of
    all three splits, whereas for DTW-KNN distances were computed only
    once).
  }
  \begin{tabular}{lrrrr}
    \toprule
    \textbf{Method}  & Raw-TCN          & MGP-RNN          & MGP-TCN          & DTW-KNN\\
    \textbf{Runtime} & \SI{49.7}{\hour} & \SI{74.8}{\hour} & \SI{73.4}{\hour} & \SI{136.9}{\hour}\\
    \bottomrule
  \end{tabular}
  \label{Supp:Runtimes}
\end{table}

\bibliographystyleappendix{plain}
\bibliographyappendix{main}

\end{document}